\documentclass[letterpaper, 10 pt, conference]{ieeeconf}  
\IEEEoverridecommandlockouts                              
\usepackage{graphicx}
\usepackage{epsfig} 
\usepackage{mathptmx}
\usepackage{mathtools}
\usepackage{amsmath} 
\usepackage{bm}

\usepackage{amssymb}  
\usepackage{amsthm}
\usepackage{siunitx}
\usepackage{wasysym}
\usepackage{wasysym}
\usepackage[mathcal]{euscript}
\usepackage{bbm}
\usepackage{color}
\usepackage{lipsum}
\usepackage{booktabs}
\usepackage[ruled,vlined,linesnumbered]{algorithm2e}\usepackage[colorlinks=true,linkcolor=black,anchorcolor=black,citecolor=black,filecolor=black,menucolor=black,runcolor=black,urlcolor=black]{hyperref}
\usepackage{etoolbox}
\usepackage[table,xcdraw,dvipsnames]{xcolor}
\usepackage{multirow}
\usepackage{bbm}
\usepackage{outlines}
\let\labelindent\relax
\usepackage{enumitem}
\setenumerate[2]{label=\alph*.}
\setenumerate[3]{label=\roman*.}

\makeatletter
\patchcmd{\@makecaption}
  {\scshape}
  {}
  {}
  {}
\makeatother

\usepackage[
    style=ieee,
    doi=false,
    isbn=false,
    url=false,
    eprint=false,
    backend=biber,
    natbib=true,
    citestyle=numeric-comp
    ]{biblatex}

\usepackage{caption}
\bibliography{references}

\usepackage{dirtytalk}

\usepackage[capitalise]{cleveref}
\crefname{equation}{}{}
\crefname{figure}{Fig.}{Figs.}
\crefname{tabular}{Tab.}{Tabs.}
\crefname{section}{Sec.}{Secs.}

\newtheorem*{problem*}{Problem}

\providecommand{\R}{\ensuremath \mathbb{R}}
\providecommand{\N}{\ensuremath \mathbb{N}}

\newcommand{\natlang}{\mathbb{L}}

\newcommand{\regtext}[1]{\mathrm{\textnormal{#1}}}

\newcommand{\tss}[1]{\textsuperscript{#1}}

\newcommand{\lbl}[1]{_{\regtext{#1}}}

\newcommand{\norm}[1]{\left\Vert#1\right\Vert}

\newcommand{\normaldist}{\mc{N}}
\newcommand{\mean}{\mu}
\newcommand{\Std}{\Sigma}
\newcommand{\std}{\sigma}
\newcommand{\given}{\mid}

\newcommand{\mc}[1]{\mathcal{#1}}
\newcommand{\pred}{\lbl{pred}}
\newcommand{\obs}{\lbl{obs}}
\DeclareMathOperator{\llm}{LLM}
\DeclareMathOperator{\stt}{STT}
\DeclareMathOperator{\tcn}{TCN}
\DeclareMathOperator{\GoalGMM}{GoalGMM}

\newcommand{\goal}{\mathbf{g}}
\newcommand{\position}{\mathbf{x}}
\newcommand{\latent}{\mathbf{h}}

\title{\LARGE \bf
Language Conditioning Improves Accuracy \\ of Aircraft Goal Prediction in Non-Towered Airspace
}
\author{
Sundhar Vinodh Sangeetha\tss{1},
Chih-Yuan Chiu\tss{2},
Sarah H.Q. Li\tss{1,$\dagger$},
Shreyas Kousik\tss{3,$\dagger$}
\thanks{
All authors are with the Georgia Institute of Technology, Atlanta, GA, USA.
\tss{1} School of Aerospace Engineering.
\tss{2} School of Electrical and Computer Engineering.
\tss{3} School of Mechanical Engineering.
$\dagger$ indicates equal advising.
Corresponding author: \texttt{ssangeetha3@gatech.edu}.
\raggedright\textbf{Codebase}: 
\href{https://github.com/sundharvs/language-conditioned-aircraft-goal-prediction}{https://github.com/sundharvs/language-conditioned-aircraft-goal-prediction}
}
}

\begin{document}

\maketitle

\begin{abstract}
Autonomous aircraft must safely operate in non-towered airspace, where coordination relies on voice-based communication among human pilots.
Safe operation requires an aircraft to predict the intent, and corresponding goal location, of other aircraft.
This paper introduces a multimodal framework for aircraft goal prediction that integrates natural language understanding with spatial reasoning to improve autonomous decision-making in such environments.
We leverage automatic speech recognition and large language models to transcribe and interpret pilot radio calls, identify aircraft, and extract discrete intent labels.
These intent labels are fused with observed trajectories to condition a temporal convolutional network and Gaussian mixture model for probabilistic goal prediction.
Our method significantly reduces goal prediction error compared to baselines that rely solely on motion history, demonstrating that language-conditioned prediction increases prediction accuracy.
Experiments on a real-world dataset from a non-towered airport validate the approach and highlight its potential to enable socially aware, language-conditioned robotic motion planning.
\end{abstract}

\section{Introduction} \label{sec:intro}

In recent years, there has been a surge of industry interest in operating autonomous aircraft from small, regional airports that operate without traffic control towers.
These non-towered airports account for $92\%$ of airports within the United States and $90\%$ of airports worldwide \cite{FAA2024AirTrafficByTheNumbers}.
Companies such as Reliable Robotics, Merlin Labs, and Ribbit have been actively developing systems for remotely piloted and fully autonomous aircraft, driven by the potential to reduce operational costs and expand access~\cite{wang2025flight}.
Despite the surging interest, safe integration of autonomous aircraft operations near non-towered airports remains incredibly challenging. 
This is in large part because non-towered airports lack towered airports' centralized surveillance, infrastructure, and air traffic controllers to ensure flight safety and resolve real-time flight path conflicts. 
Instead, potential conflicts are locally managed by pilots through a combination of visual flight rules and self-announced radio calls over the common traffic advisory frequency (CTAF), enabling pilots to detect and avoid one another.
Therefore, flight safety near non-towered airports requires not only adherence to established aviation rules, but also the ability to interpret loosely structured natural language communications in real time.

However, current state-of-the-art conflict avoidance approaches for autonomous aircraft strictly rely only on observed trajectories and structured air traffic rules~\cite{yang2021autonomous}, and do not consider radio communication between pilots, to the best of our knowledge. 
This means that, near non-towered airports, current methods will miss critical coordination information. %
This makes mixed human-autonomy operations dangerous, and potentially requires human pilots to accommodate and increase safety margins around autonomous aircraft.
Towards solving this challenge, this paper takes key steps towards enabling robot aircraft to understand and act on human radio communication.

\begin{figure}[t]
    \centering
    \includegraphics[width=0.95\columnwidth]{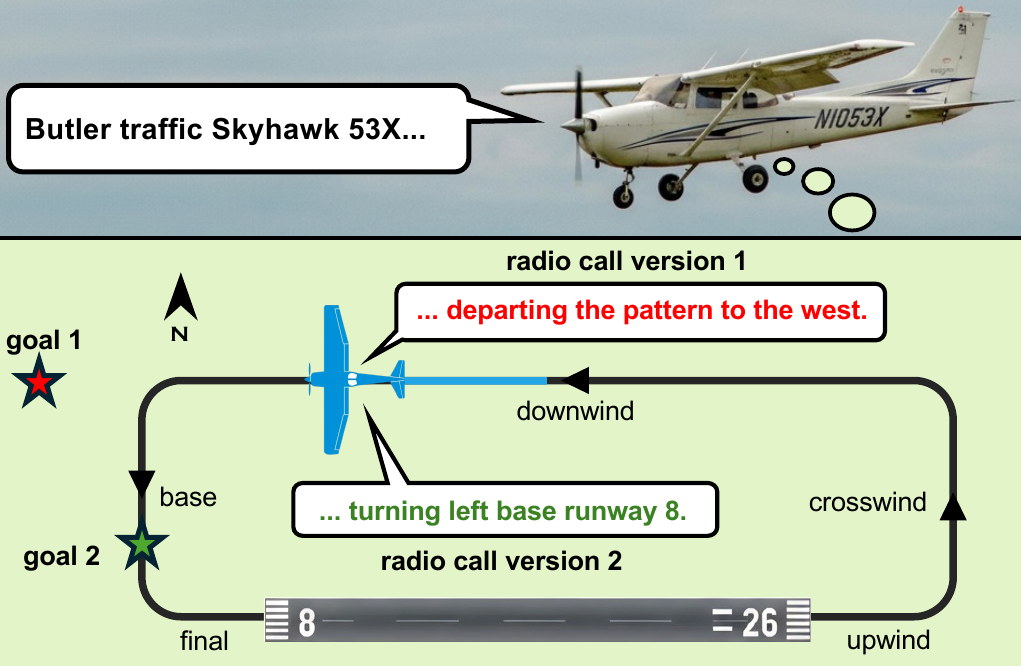}
    \caption{Example of a pilot radio call on the CTAF in non-towered airspace.
    An aircraft (blue in inset) makes a radio call announcing its intent with respect to the ``Runway 8'' left traffic pattern.
    Multiple goals are possible depending on the radio call.
    In this work, we show that conditioning on such radio calls along with the aircraft's recent trajectory (blue line on downwind leg) improves accuracy of goal prediction, which is critical for autonomous operation in non-towered airspace.}
    \label{fig:front_figure}
    \vspace*{-15pt}
\end{figure}

In this work, we aim to bridge the gap between unstructured human pilot communication and autonomous collision avoidance.
That is, this work makes a key step towards addressing the broad question,  how should an autonomous aircraft interpret and use radio communications to safely and effectively coordinate with other (potentially human-piloted) aircraft in a non-towered airport environment?
This challenge can be broken into two research problems: predicting the motion of other aircraft using radio messages, and announcing the ego motion plan to other aircraft in a timely way while obeying radio communication rules.
In this paper, we focus on the first problem by answering the following question in the affirmative: does incorporating language conditioning improve a robot aircraft's ability to predict the motion of other aircraft in its vicinity?

\subsection{Related Work}
Existing research in decentralized air traffic control primarily relies on historical observations to predict future trajectories~\cite{6644284}.
Many works have recognized the importance of predicting intent to forecast future aircraft trajectories \cite{yepes2007new,yin2025aircraft,mohamed2020social,giuliari2021transformer,patrikar2022predicting,navarro2022social,yang2025goodflight,patrikar2022predicting,tran2022aircraft,pang2020conditional,wang2021voice,pang2025voice,guo2024integrating,rakas2024controller}.
Generally, they approximate pilot intent as a function of the aircraft's past trajectory.
Foundational methods in this area explicitly model aircraft dynamics and use traditional statistical methods like Kalman filters and maximum likelihood inference to predict future trajectories \cite{yepes2007new}.
Recent works have instead used deep learning approaches to build intent-driven prediction models with increasing complexity.
For example, these models use additional information like maneuvering patterns from a repository of historical trajectories around an airport \cite{yin2025aircraft} and wind data \cite{patrikar2022predicting}.
They also employ advanced model architectures that have increased predictive power, such as encoder-decoder architecture \cite{tran2022aircraft}, generative adversarial networks \cite{pang2020conditional}, and diffusion models \cite{yang2025goodflight}.
Some recent works have recognized that aircraft radio communications contain intent information and developed trajectory prediction models conditioned on these communications \cite{wang2021voice,pang2025voice,guo2024integrating,rakas2024controller}.
However, these works consider direct instructions from air traffic controllers to aircraft, which map clearly to trajectory parameters like speed, altitude, and GPS waypoints.
Instead, we consider non-towered airport communications, which are less structured.

We note that the robotics community at large is experiencing rapid advances in integrating natural language processing with perception and decision-making \cite{wang2025large,openvla,kim2024survey}.
Recent developments in speech-to-text (STT) and large language models (LLMs) have enabled robots to interpret audio messages and leverage linguistic information for reasoning about their environment.
Within this broader trend, our work seeks to demonstrate how natural language understanding can be effectively applied to the spatial trajectory prediction problem, thereby bridging the gap between unstructured human communication and structured motion planning.

\subsection{Contributions}
We make three contributions:
\begin{enumerate}
    \item an automatic speech recognition method to transcribe radio calls on CTAF and identify speakers,
    \item a method to infer intent from radio calls in uncontrolled airspace and predict aircraft trajectories with reduced error, and
    \item an experimental evaluation showing how language conditioning improves goal prediction performance on real-world non-towered airport data.
\end{enumerate}
\section{Preliminaries and Problem Formulation} \label{sec:problem_statement}
Consider an autonomous aircraft operating in the airspace surrounding an uncontrolled airport, in which human-piloted aircraft are also operating.
How can this aircraft forecast the future goals of other aircraft in this airspace using observations of their past trajectories as well as their natural language radio transmissions on the CTAF?
We focus on predicting goals because it has been shown that accurate goal prediction greatly improves the quality of downstream \textit{trajectory} prediction for motion planning \cite{mangalam2020not,choi2021drogon,yang2025goodflight,li2023conditional,fadillah2025goalnet}.

Throughout this paper, we consider the prediction problem for a single aircraft, without conditioning on information about other aircraft.
We note that other settings (e.g., autonomous cars) that do \textit{not} have radio communication \textit{do} rely on multi-agent conditioning \cite{salzmann2020trajectron}.
However, since radio calls come from human pilots who are aware of each other socially, we assume that these calls implicitly encode potential interactions, and therefore we focus on predicting individual agents' goals, leaving extensions to multi-agent prediction for future work.

To proceed, we first introduce our notation conventions, then formalize our problem statement, and finally briefly discuss details relevant to non-towered airspace.

\subsection{Notation}
Let $\R$ denote the reals and $\N$ the natural numbers.
Let $\natlang$ denote the space of all possible natural language utterances.
Consider an aircraft whose intent and goal we wish to predict.
Let $\position(t) = \big(x(t), y(t), z(t)\big)$ denote its physical position at a time $t \in \R$.
Let $r(t) \in \natlang$ denote the natural language radio call made by the aircraft at time $t$.
We use $\mc{T} = \{t_0,\dots,t\obs\} \subset \R$ to denote the observation time horizon, such that we observe a trajectory $\big(\position(t)\big)_{t \in \mc{T}}$ of the aircraft\footnote{This information is publicly available from aircraft transponder data.}.
Let $t_r \leq t\obs$ denote the time of the aircraft's latest radio call, and a time instance $t\pred  > t\obs$ to denote the chosen time step for goal prediction for all aircraft.

\subsection{Problem Statement and Solution Overview}
The technical problem we seek to solve is how to predict an aircraft's intent, represented as a goal position, given its recent trajectory and radio calls:
\begin{problem*}
Suppose we have observed an aircraft's recent trajectory $\big(\position(t)\big)_{t \in \mc{T}}$ and latest radio call $r(t_r) \in \natlang$.
We seek to predict the aircraft's future goal $\goal$, meaning its position at $t\pred > t\obs$, denoted
\begin{equation}
\label{eqn:future_goal}
    \goal = \position(t\pred).
\end{equation}
\end{problem*}

To solve this problem, we model a probability distribution over the aircraft's future goal position, conditioned on its observed trajectories and latest radio call, and approximate its $\goal$ as a sample $\hat{\goal}$ drawn from this distribution:
\begin{equation}\label{eqn:solution_distribution}
     \goal \approx
     \hat{\goal}
     \sim
     p\left(\hat{\goal} \mid
            \big(\position(t)\big)_{t \in \mc{T}}, r(t_r)
     \right).
\end{equation}

\subsection{Characterizing Non-Towered Terminal Airspace}
As standardized by the FAA~\cite{FAA-AC-90-66B}, for all non-towered airports in the US, the airspace surrounding the airport is structured around the traffic pattern.
As shown in Figure \ref{fig:front_figure}, this terminal airspace is divided into upwind, crosswind, downwind, base, and final legs of the traffic pattern for each runway.
Active runways also adopt left-handed or right-handed traffic conventions.
For example, ``left downwind'' refers to the segment of the traffic pattern reached by making two left-hand $90^\circ$ turns after departure, and ``right downwind'' refers to the segment reached by making two right-hand $90^\circ$ turns after departure.
Unless otherwise specified, we use the left-handed traffic pattern convention in the following sections, which is the default convention for non-towered airports.

Radio calls on CTAF frequencies in non-towered airspace typically contain four pieces of information: airport name, aircraft identifier, position, and intention.
For example, ``Butler County traffic Cherokee 135PL three miles south of the field entering left downwind runway 8.'' 
Reported positions and intentions are typically a combination of distance, cardinal direction, and traffic pattern segment.
Critically, pilots often abbreviate or omit some information, present it in various orders, and potentially insert verbal filler, all of which necessitates semantic understanding of radio calls.
\section{Method}\label{sec:method}

\begin{figure*}[t]
    \centering
    \includegraphics[width=0.8\textwidth]{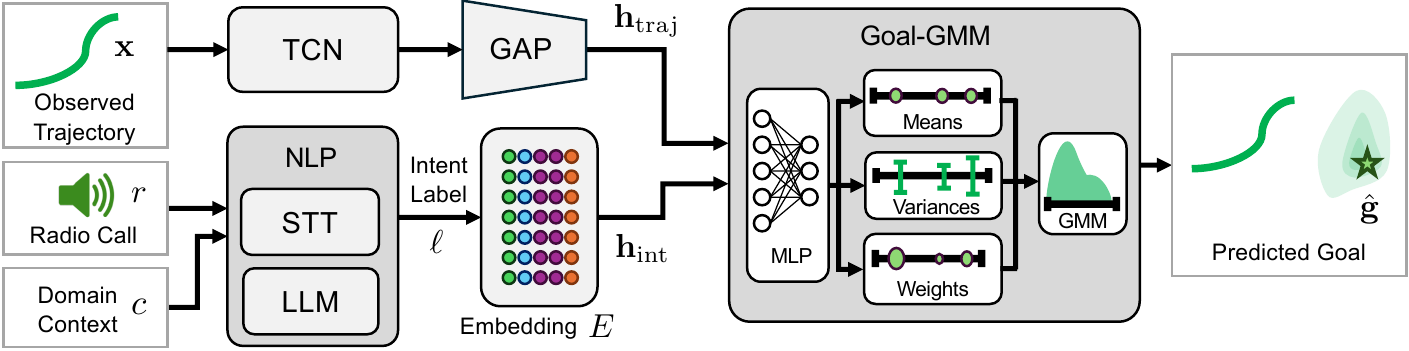}
    \caption{An overview of our model architecture as described in \Cref{subsec: goal prediction}.
    Our key finding is that conditioning on natural language radio calls improves autonomous aircraft's ability to predict the intent of other aircraft in non-towered airspace.}
    \label{fig: overview}
    \vspace*{-15pt}
\end{figure*}

In this section, we present our framework for predicting aircraft goals (i.e., trajectory endpoints) in non-towered airspace by conditioning on the observed past trajectories and natural language radio calls.
Our approach consists of three components: (\Cref{subsec: aircraft recognition via asr}) speech-to-text transcription and speaker aircraft identification to transform CTAF audio into text, (\Cref{subsec: intent modeling}) intent extraction using large language models with domain-specific context to reduce text into a compact intent representation, and (\Cref{subsec: goal prediction}) goal prediction with a learned, multimodal model, outputting a probability distribution of an aircraft goal. 

\subsection{Context Enhanced Speech-to-Text and Aircraft Identification}\label{subsec: aircraft recognition via asr}

To incorporate pilot radio communications into aircraft goal prediction, it is essential to accurately transcribe audio messages and identify the transmitting aircraft.
While trained speech-to-text ASR models exist for aviation radio communications \cite{badrinath2022automatic,subramanian2018custom,rakas2024controller,wlv3-atco2-asr}, we find that these models perform poorly on non-towered airspace radio calls, since they are trained on communications between air traffic controllers and commercial aircraft in controlled airspace \cite{zuluaga2022atco2}.
Additionally, large general purpose speech-to-text models like Whisper \cite{whisper} also do not perform well in transcribing radio calls due to audio noise and domain-specific terminology.
However, we find that by providing large general purpose models with domain-specific information, we can accurately transcribe radio calls and identify transmitting aircraft.

Specifically, in our framework, a speech-to-text model and large language model are provided with \textit{static context} $s \in \natlang$ in the form of common terminology used in non-towered airspace, runway numbers, airport name, etc., plus 50 few-shot human-labeled examples of radio calls with corresponding intents.
With each radio call, the models are also provided with \textit{dynamic context} $d \in \natlang$ in the form of a list of aircraft ADS-B identifiers (tail numbers) in the airspace surrounding the airport, their corresponding manufacturer and model names, which are retrieved from the FAA N-Number database, and their distance and cardinal direction from the airport.
This \textit{dynamic context} is automatically generated at inference time.
We define the total \textit{domain context} as the combined static and dynamic contexts:
\begin{align}
    c = (s, d) \in \natlang.
\end{align}
We provide an example dynamic context in \Cref{appx:domain_context}; the static context is too large to print due to page limits but will be included in our open-source codebase.

The dynamic context is especially crucial in accurately identifying which aircraft transmitted a radio message.
Commercial aircraft in controlled airspace must identify themselves in every radio call with the airline name and full flight number, which map directly to ADS-B identifiers.
However, general aviation aircraft, which are the ones that largely operate at non-towered airports, have greater flexibility in how they identify themselves; using combinations of manufacturer names, model names and parts of their tail number.
``N123AB'', ``Cessna 23AB'', and ``Skyhawk'' are all examples of radio call identifiers for the same aircraft, whereas ``N123AB'' is the ADS-B identifier.
Radio identification on CTAF is oriented towards visual identification of aircraft by human pilots, so identifiers like model names are most often used rather than full tail numbers.

\subsection{Intent Modeling and Extraction}\label{subsec: intent modeling}

Because pilot radio communications are inherently unstructured, directly incorporating a radio call into trajectory prediction models greatly expands the input space and complicates training.
To address this challenge, we transform radio calls into a compact form that captures the pilot's intent, which we then use as an additional input to the goal prediction mechanism.
This intent model aims to be more compact and uniform than loosely structured language sequences while remaining sufficiently rich to distinguish between different future spatial goals.

For a given aircraft, we consider its most recent radio call $r(t_r) \in \natlang$ occurring at time $t_r$.
If there has been no radio call for the past 10 minutes, we treat the radio call, and corresponding intent label as ``unknown''.

We use a discrete set of intent labels  to characterize the radio call $r(t_r)$. 
These labels specify distinct spatial approaches and locations with respect to the runway's traffic pattern as shown in \Cref{fig:front_figure}.
Specifically, the intent labels captures (a) take off, landing, and entering three traffic pattern legs (crosswind, base, downwind) for each runway end (eg. runway 8 or 26 depending on the direction used), and (b) the four different cardinal directions of leaving the overall traffic pattern (e.g. NESW).
Two additional labels, ``insufficient information" and ``other intent'', capture cases in which there is either insufficient information to determine intent or in which there is sufficient information, but the intent does not belong to one of the other classes.
Empirically, we find that the ``other intent'' label predominantly captures ground operations that do not conflict with in-air operations.

Formally, let $\mc{I} \subset \natlang$ denote the finite set of discrete intent labels.
For each radio call $r(t_r)$ from the aircraft, we extract the corresponding intent label by passing the radio call through a speech-to-text (STT) model, then an LLM conditioned on the domain context:
\begin{align}\label{eqn:intent_label}
    \ell = \llm\Big(\stt\big(r(t_r)\big) \mid c\Big) \in \mc{I}.
\end{align}
We use gpt-4o-transcribe for STT and Gemma 3 27B \cite{team2024gemma} as the LLM.

\subsection{Aircraft Goal Prediction}\label{subsec: goal prediction}

Given the observed trajectory of an aircraft and its language conditioned intent label $\ell$, our objective is to approximate the probability distribution as in \eqref{eqn:solution_distribution} over the predicted future goal $\hat{\goal}$ of the aircraft.
We approximate this distribution using a Gaussian Mixture Model (GMM), meaning a weighted sum of $K$ independent Gaussian components, as
\begin{align}\label{eqn:goalGMM_definition}
    p\Big(\hat{\goal} \mid
            \big(\position(t)\big)_{t \in \mc{T}}, r(t_r)
     \Big)
    \approx
    \sum_{k=1}^K
        \pi_k \cdot \normaldist\left(
            \mean_k,\Std_k \given
            \latent\lbl{traj},\ \latent\lbl{int}\right),
\end{align}
where we condition the distribution on latent vectors $\latent\lbl{traj}$ and $\latent\lbl{int}$ explained below.
Each $\pi_k \in [0,1]$ is the mixture weight of the $k$\tss{th} component such that $\sum_{k=1}^K \pi_k = 1$.
Each $\normaldist(\cdot)$ is a multivariate Gaussian distribution with mean $\mathbf{\mean}_k \in \R^3$ and covariance $\Std_k \in \R^3$.
We assume a diagonal covariance matrix, such that $\Std_k = \text{diag}\big((\mathbf{\std}_k)^2\big)$ for a vector $\std_k \in \R^3$.

To implement \eqref{eqn:goalGMM_definition}, as shown in Figure~\ref{fig: overview}, our proposed architecture processes the observed trajectory and intent label in parallel before combining them to predict the GMM parameters in the ``GoalGMM'' component.
GoalGMM itself consists of three linear heads that output the GMM parameters: mean, variance, and mixture weight.
Next, we detail the trajectory and intent processing and each prediction head.

\subsubsection*{Trajectory Encoder}
To capture the motion patterns and temporal dependencies within the observed trajectory, we employ a Temporal Convolutional Network (TCN) \cite{lea2016tcn}.
The TCN processes each input trajectory $\big(\position(t)\big)_{t \in \mc{T}}$ using a stack of causal, dilated, 1-D convolutional layers with channel sizes of 64, 128, and 256.
The TCN maps the input trajectory sequence to a sequence of high-dimensional feature vectors
\begin{align}
    \big(\latent(t_0), \dots, \latent(t\obs)\big) = \tcn\big((\position(t))_{t \in \mc{T}}\big),
\end{align}
where each $\latent(t) \in \R^{n\lbl{traj}}$, where $n\lbl{traj} = 256$ in our implementation.
To produce a single, fixed-size representation, we apply global average pooling (GAP) over the time dimension:
\begin{equation}
    \latent\lbl{traj} = \frac{1}{|\mc{T}|} \sum_{t \in\mc{T}} \latent(t).
\end{equation}
The resulting vector $\latent\lbl{traj}$ serves as a summary of the aircraft's observed trajectory.

\subsubsection*{Intent Embedding}

To incorporate intent information from radio calls, the discrete intent label $\ell \in \mc{I}$ is mapped to a dense vector representation using a learned embedding layer.
This allows the model to learn a meaningful, continuous representation for each of the $|\mc{I}|$ possible intents.
The embedding for the aircraft is produced as:
\begin{equation}
    \latent\lbl{int} = E(\ell),
\end{equation}
where $E: \mc{I} \to \R^{n\lbl{int}}$ is a learnable embedding dictionary and $\latent\lbl{int} \in \R^{n\lbl{int}}$ is the resulting intent vector.
In our implementation, $n\lbl{int} = 64$.

\subsubsection*{Probabilistic Prediction Head}
These latent vectors $\latent\lbl{traj}$ and $\latent\lbl{int}$ are concatenated and passed through a shared Multi-Layer Perceptron (MLP) with hidden dimensions of $[128, 64]$.
This representation is then fed in parallel to three distinct linear heads to predict the parameters of a GMM with $K$ components:
\begin{enumerate}
    \item \textit{Mean Head}: Predicts the means $\mathbf{\mean}_k \in \R^{K \times 3}$ for each of the $K$ Gaussian components.
    \item \textit{Variance Head}: Predicts variances as log-variances $\log(\mathbf{\std}^2)_k \in \R^{K \times 3}$ for numerical stability.
    \item \textit{Mixture Weight Head}: Predicts the logits for the mixing coefficients $\pi_k \in [0,1]$, which determine the weight of each component.
\end{enumerate}
In our implementation, $K=15$.
Putting these parts together, we have
\begin{align}
    \left\{\mean_k,\Std_k,\pi_k\right\}_{k=1}^K = \GoalGMM(\latent\lbl{traj}, \latent\lbl{int}).
\end{align}
Then, the model is trained end-to-end by minimizing the negative log-likelihood loss $\mc{L}$ of the ground-truth goal under the predicted GMM distribution, plus an entropy loss encouraging the GMM modes to diverge
\begin{align}
    \mc{L} &= -\frac{1}{N}\sum_{i=1}^N \log
        \left(
            \sum_{k=1}^K \pi_k\cdot f_{\normaldist}(\goal_i ; \mean_k,\Std_k)
        \right)+ \mc{L}\lbl{ent}
\end{align}
where $f_\normaldist(x; \mean,\Std)$ denotes the PDF of a normal distribution with mean $\mu$ and covariance $\Std$ evaluated at $x$, $N$ is the number of goals per batch, and $\mc{L}\lbl{ent}$ is as in \cite[Eq. (7)]{zhou2020movement}.

Training is performed for 150 epochs using the AdamW optimizer.
The learning rate is initialized at $1 \times 10^{-4}$ and reduced by a factor of 0.2 after 5 epochs without improvement in training loss.

\begin{figure}[t]
    \centering
    \includegraphics[width=\linewidth]{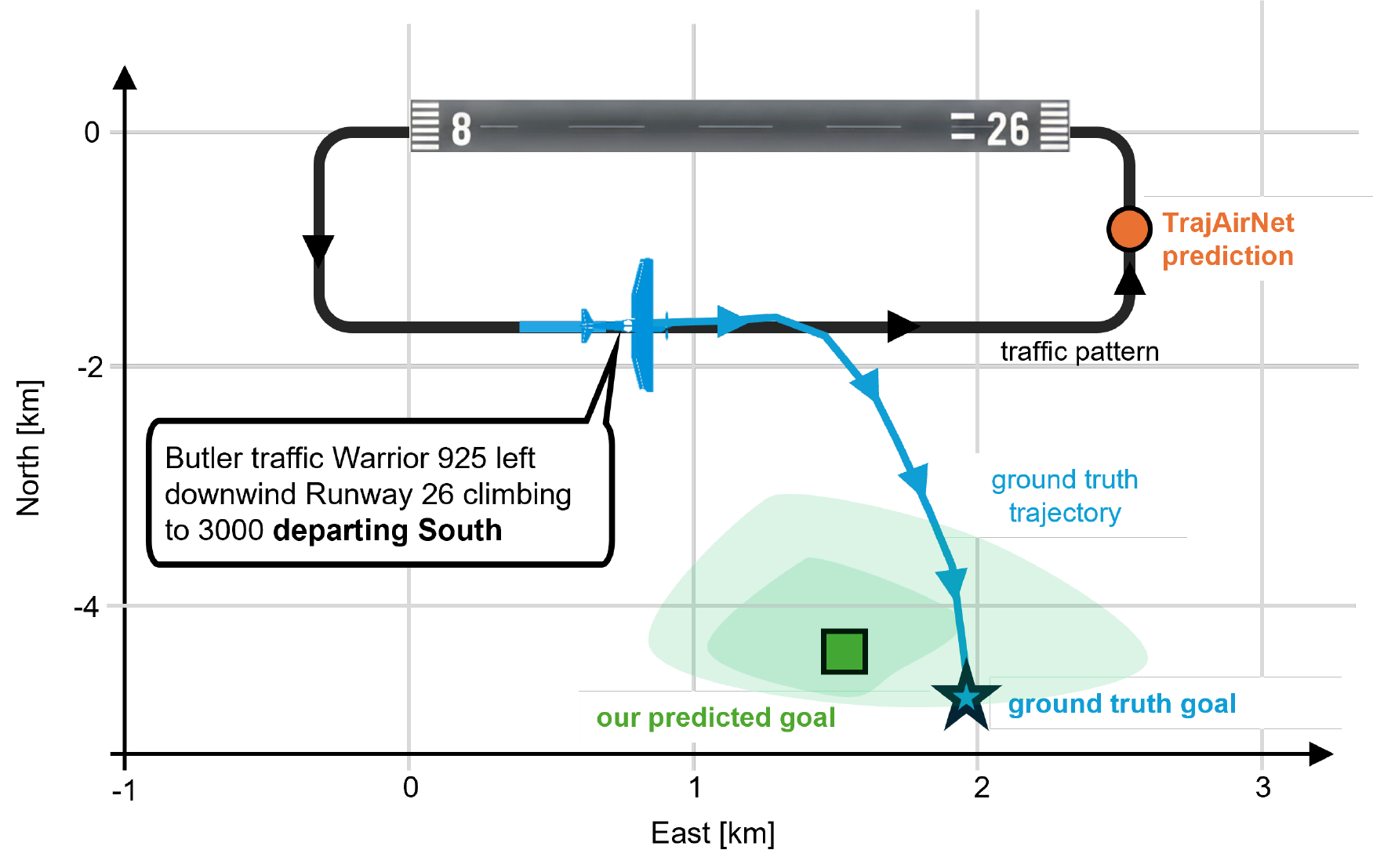}
    \caption{Example trajectory prediction comparing our method to TrajAirNet \cite{patrikar2022predicting}, showing the improvement from using radio call info (in speech bubble).}
    \label{fig:comparison}
    \vspace*{-15pt}
\end{figure}
\section{Experiments} \label{sec:experiments}
We now evaluate our method by testing speech recognition accuracy (\Cref{subsec: ASR experiment}), language-conditioned goal prediction (our core experiment, \Cref{subsec: goal prediction}), and feature/language ablations (\Cref{subsec: feature importance experiment,subsec: varying_information}).
All experiments are conducted on a workstation with a 32-core AMD Ryzen Threadripper PRO 7955WX CPU, 256 GB RAM, and an NVIDIA RTX 6000 Ada Generation GPU (48 GB VRAM).

\subsection{Automatic Speech Recognition (ASR)}\label{subsec: ASR experiment}

First, we evaluate the accuracy of our automatic speech recognition approach on transcribing historical radio communications on the CTAF observed at an uncontrolled airport, identifying the ADS-B IDs of aircraft that transmit messages, and labeling their intent from their messages.

\subsubsection*{Hypothesis}
We hypothesize that given sufficient domain context\footnote{Domain context is defined in \cref{subsec: aircraft recognition via asr}. An example is provided in \cref{appx:domain_context}}, general purpose ASR models can transcribe aviation radio communications well without any finetuning, and that large language models can identify which aircraft transmitted a radio call and accurately label its intent.

\subsubsection*{Experiment Design}
We ran the experiments on a dataset of 250 radio call audios from the TartanAviation dataset \cite{tartanaviation} using gpt-4o-transcribe for speech-to-text and Gemma 3 27B for speaker identification and intent inference.
The performance of our method is measured by comparing against human transcribed and annotated radio calls.

\subsubsection*{Metrics}
We report the word error rate (WER) of our method, which is a common metric for ASR \cite{fiscus1997post}, the percentage of correctly identified speaker aircraft, and the percentage of correctly identified intent labels.

\begin{table}[ht]
\centering
\resizebox{\linewidth}{!}{
\begin{tabular}{ccccc}
\toprule
\textbf{Method} & \textbf{WER} & \textbf{SIA (\%)} & \textbf{ILA (\%)} \\
\midrule
With domain context     &  \textbf{33.97}  &  \textbf{94.8}   &  \textbf{82.8}  \\
Without domain context      &    60.55     &  63.6  & 32.8  \\
\bottomrule
\end{tabular}
}
\caption{ASR accuracy of aviation radio communications in uncontrolled airspace. SIA denotes speaker identification accuracy, ILA denotes intent labeling accuracy. Including domain-specific context significantly improves speech-to-text, speaker identification, and intent labeling performance.}
\label{tab:asr}
\end{table}

\subsubsection*{Results and Discussion}
The results of this experiment are summarized in Table \ref{tab:asr}.
We observe that including domain context through gpt-4o-transcribe and Gemma 3 27B prompt input improves transcription, speaker identification, and intent labeling accuracy.
It has been noted in the automatic speech recognition literature that minor differences in transcripts can lead to large word error rates despite accurately capturing meaning \cite{whisper}.
Our results support this notion and suggest that these errors do not have a large impact on the ability of LLMs to accurately extract meaning from text.
Despite relatively large word error rates, we achieve high accuracy in speaker identification and intent labeling from transcription.

\subsection{Language Conditioned Goal Prediction}
\label{subsec: goal prediction experiment}

In this experiment, we test our goal prediction model conditioned on aircraft intent labels extracted from transcribed CTAF radio communications.
We feed real-world aircraft trajectory data and radio call transcripts into our prediction model and compare its outputs against ground-truth trajectories.

\subsubsection*{Hypothesis}
We hypothesize that conditioning the aircraft goal prediction model on intent labels from radio calls will decrease final displacement error (FDE) of predicted future goals compared to only conditioning the model on the position histories and wind contexts of each aircraft.

\subsubsection*{Experiment Design}
We tested on aircraft trajectory and audio data collected at the Pittsburgh-Butler Regional Airport (KBTP) from the TartanAviation dataset \cite{tartanaviation}.
A number of works \cite{patrikar2022predicting,navarro2022social,yang2025goodflight} test their methods on a subset of this data made available in an earlier publication, the TrajAir dataset \cite{patrikar2022predicting}, labeled as ``7days-1'', ``7days-2'', ``7days-3'', and ``7days-4''.
However, audio files for many time periods, including entire days, are missing from this subset.
Importantly, ``7days-4'' does not have any audio data and ``7days-1'' only has one day of audio data, so we do not report results for these subsets.
To compare our method fairly to these previous works, we evaluate on the subset containing audio.
Additionally, we conduct our other experiments on a one week subset from the larger TartanAviation dataset (labeled 7daysJune) that contains complete audio and trajectory data.

As in TrajAirNet \cite{patrikar2022predicting}, the observation time horizon is set to 11 seconds, and the prediction time horizon is set to 120 seconds.
We study the impact of changing these parameters in \cref{subsec: varying_information}.

\subsubsection*{Metrics}
We use a standard metric for trajectory prediction to quantify the performance of our model: best-of-$N$ (with $N = 10$) final displacement error (FDE) \cite{zhao2020noticing}, defined as the average over all aircraft of the lowest error,
\begin{align}
    \text{FDE} = \frac{1}{|\mc{A}|} \sum_{a\in\mc{A}} \min_{j \in \{1,\dots,N\}} \norm{
        \goal_{a,j} - \hat{\goal}_{a,j}
    }_2,
\end{align}
where $\mc{A} = \{1,\cdots,n\lbl{agt}\}$ indexes all aircraft in the airspace.
This metric is chosen to allow comparison of trajectory prediction performance to a large number of existing methods by averaging across all goal positions predicted across an entire test dataset.

However, FDE alone does not capture the variance in the prediction performance across different testing sets, which is critical to understand a method's precision for downstream motion planning; that is, if a method has high variance, then it is more difficult to use for planning collision-avoiding motions, even if it is accurate.
Thus, as measures of dispersion in FDE, we report standard deviation in tables and show inter-quartile range (IQR) (to visualize skew) in figures.

\subsubsection*{Baselines}
The main baseline we compare to is TrajAirNet~\cite{patrikar2022predicting}, which has open source code; that work was one of the first to consider this specific problem, and introduced constant velocity and nearest neighbor baselines.
Following the literature, we also report results claimed by several baselines directly from the corresponding papers \cite{mohamed2020social,giuliari2021transformer,patrikar2022predicting,navarro2022social,yang2025goodflight}, as code was either not available or could not be run without errors; we are working on reproducing these methods' results for future work.
We further note that these do not report any variance or dispersion measures in their FDE, which makes it difficult to understand or compare against how precise these methods are.
To handle this, we rerun TrajAirNet to confirm their results and compute variance.

\begin{table}[htbp]
\centering
\resizebox{\columnwidth}{!}{
\begin{tabular}{lcccc}
\toprule
\textbf{Algorithm} & \textbf{7Days-1} & \textbf{7Days-2} & \textbf{7Days-3} & \textbf{7Days-4} \\
\midrule
Const. Vel. \cite{salzmann2020trajectron} c.f. \cite{patrikar2022predicting} & 4.08 & 4.31 & 4.30 & 4.16 \\
Nearest Neigh. \cite{patrikar2022predicting} & 2.70 & 1.99 & 2.69 & 2.58 \\
STG-CNN \cite{mohamed2020social} c.f. \cite{patrikar2022predicting} & 2.35 & 2.70 & 2.67 & 2.29 \\
TransformerTF \cite{giuliari2021transformer} c.f. \cite{patrikar2022predicting} & 3.85 & 4.10 & 4.36 & 4.19 \\
TrajAirNet \cite{patrikar2022predicting} & 1.42 & 1.63 & 1.72 & 1.41 \\
Social-PatteRNN-ATT \cite{navarro2022social}& 1.42 & 1.67 & 1.65 & 1.51\\
GooDFlight \cite{yang2025goodflight}& 0.41 & 0.40 & 0.48 & 0.40\\
\midrule
TrajAirNet \cite{patrikar2022predicting} (rerun) &  $1.46 \pm 0.63$ & $1.72 \pm 0.60$ & $1.91 \pm 0.61$ & $1.64 \pm 0.77$\\
\textbf{Ours} & \textbf{-} & \textbf{$0.71\pm0.81$} & \textbf{$0.69 \pm 0.57$} & \textbf{-} \\
\bottomrule
\end{tabular}
}
\caption{Goal prediction error (FDE) [km].}
\label{tab:trajectory_prediction_error}
\vspace*{-15pt}
\end{table}

\begin{table}[htbp]
\centering
\resizebox{\columnwidth}{!}{
\begin{tabular}{lcccc}
\toprule
\textbf{Method} & \textbf{Mean FDE} & \textbf{25th pctl.} & \textbf{75th pctl.} & \textbf{St. Dev.} \\
\midrule
TrajAirNet \cite{patrikar2022predicting} & 1.390 & 0.860 & 1.731 & 0.686 \\
\midrule
\textbf{Ours} & \textbf{0.486} & \textbf{0.160} & \textbf{0.578} & \textbf{0.453} \\
\bottomrule
\end{tabular}
}
\caption{Goal prediction error (FDE) [km] in the 7daysJune subset.}
\label{tab: trajairnet comparison}
\vspace*{-5pt}
\end{table}

\subsubsection*{Results and Discussion}
The results of this experiment are summarized in Table \ref{tab:trajectory_prediction_error}.
Across all datasets, we show significant improvement over all but one of the baselines, GooDFlight \cite{yang2025goodflight}, and note that the improvement in FDE falls outside the standard deviation, suggesting a statistically significant improvement.

Although we do not achieve higher performance than in GooDFlight, we note that our method to infer goals from language is complementary to the goal-conditioned model in GooDFlight, and that our primary goal is to show the improvement in goal prediction performance as a result of language conditioning.
We also note that the transformer-based diffusion model in GooDFlight has a much larger model capacity, increasing its predictive power and perhaps yielding an even larger improvement if combined with our language conditioning.
However, as the code for this model is not open-source, we are unable to test this hypothesis.

Finally, we test our model on the 7daysJune dataset with complete audio and trajectory data and report lower error as well as dispersion statistics in \cref{tab: trajairnet comparison}.
An example goal prediction is shown in \Cref{fig:comparison}.

\subsection{Feature Importance Study}
\label{subsec: feature importance experiment}

We evaluate the impact of language conditioning on goal prediction by conducting a feature importance study on the intent labels extracted from radio calls, which serve as features for our model.
Although it is difficult to map exactly how specific inputs affect model output with a deep learning model, we evaluate feature importance using two well-known methods: Permutation Feature Importance (PFI) \cite[\S10]{breiman2001random} and Leave One Feature Out (LOFO).

\subsubsection*{Hypothesis}
We hypothesize that removing intent information degrades goal prediction performance. 

\subsubsection*{Experiment Design}
In practice, evaluating PFI involves testing the pre-trained multimodal model on a dataset with shuffled aircraft intent labels.
The corresponding increase in prediction error is used to characterize the reliance of the model on intent label features.

Evaluating LOFO involves training a new model that is not conditioned on intent labels and only on aircrafts' past trajectories.
The difference in error relative to the full model characterizes importance.

\subsubsection*{Metrics}
For evaluating PFI and LOFO, we track change in the previously defined prediction error metric: FDE.

\subsubsection*{Results and Discussion}
Ablating away the language-conditioned intent labels shows the impact of this information in improving goal prediction performance.
With shuffled intent labels (PFI), we observe a modest increase in best-of-$N$ FDE of $+0.163$ compared to the baseline.
This indicates that our model makes systematic use of intent features.

With intent features entirely removed (LOFO), the degradation is much larger: sampled FDE increases by $+0.598$ on average (95\% CI $[0.591, 0.604]$, $p<0.001$).
This result confirms that intent labels provide substantial predictive value beyond trajectory information alone.

\subsection{Performance Under Varying Information Conditions}
\label{subsec: varying_information}
We observe that many previous works in trajectory prediction only present results where certain specific observation and prediction time horizons are used (c.f., \cite{patrikar2022predicting, yang2025goodflight, navarro2022social}).
Instead, we aim to study the change in our model's performance as information parameters such as these are changed.
We analyze the performance of our model by evaluating its sensitivity to several key factors:
\begin{enumerate}
    \item The observation time horizon, $t\obs$.
    \item The prediction time horizon, $t\pred$.
    \item The time difference between the prediction time and the time the most recent radio call was received.
    \item Radio calls with ``Unknown'' extracted intent labels.
\end{enumerate}
\subsubsection*{Hypothesis}
We hypothesize that our language-conditioned trajectory prediction model architecture is robust to changes in observation time horizon, prediction time horizon, and time difference between inference and radio call, and that trajectory prediction error is larger for ``Unknown'' intent labels.

\subsubsection*{Results and Discussion}

\begin{figure}[htbp]
    \centering
    \includegraphics[width=0.95\linewidth]{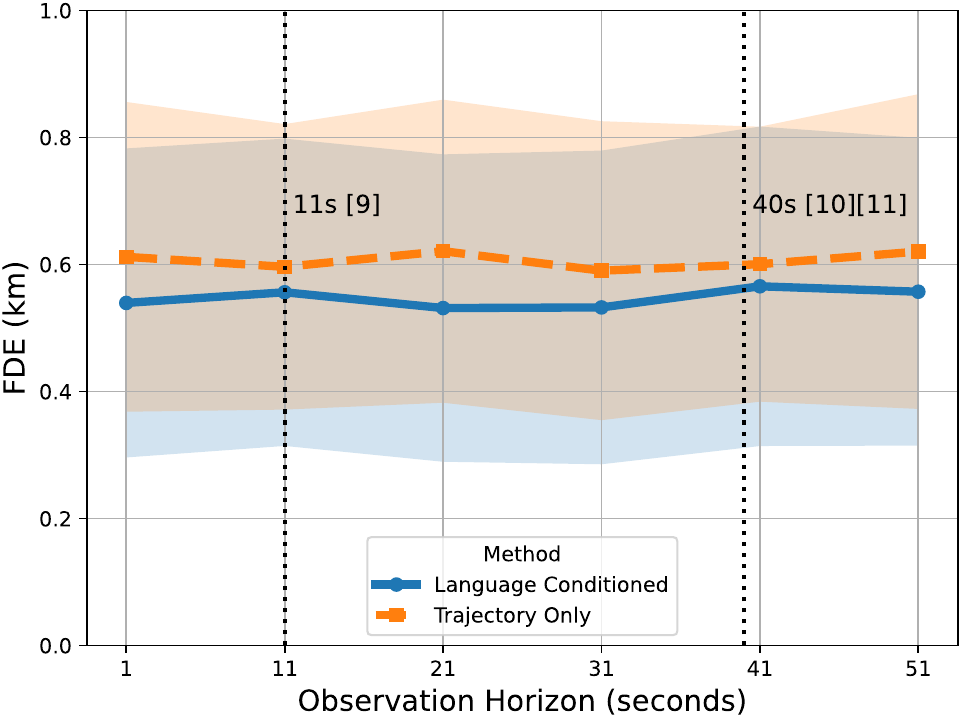}
    \caption{Final displacement error (FDE) of predicted aircraft goals with respect to observation time horizon. Longer observation windows do not improve goal prediction accuracy. The observation horizons used in TrajAirNet \cite{patrikar2022predicting}, GooDFlight \cite{yang2025goodflight} and Social-PatteRNN-ATT \cite{navarro2022social} are shown as vertical lines. IQR is plotted as a measure of dispersion.}
    \label{fig:obs_horizon}
    \vspace*{-8pt}
\end{figure}

\begin{figure}[htbp]
    \centering
    \includegraphics[width=0.95\linewidth]{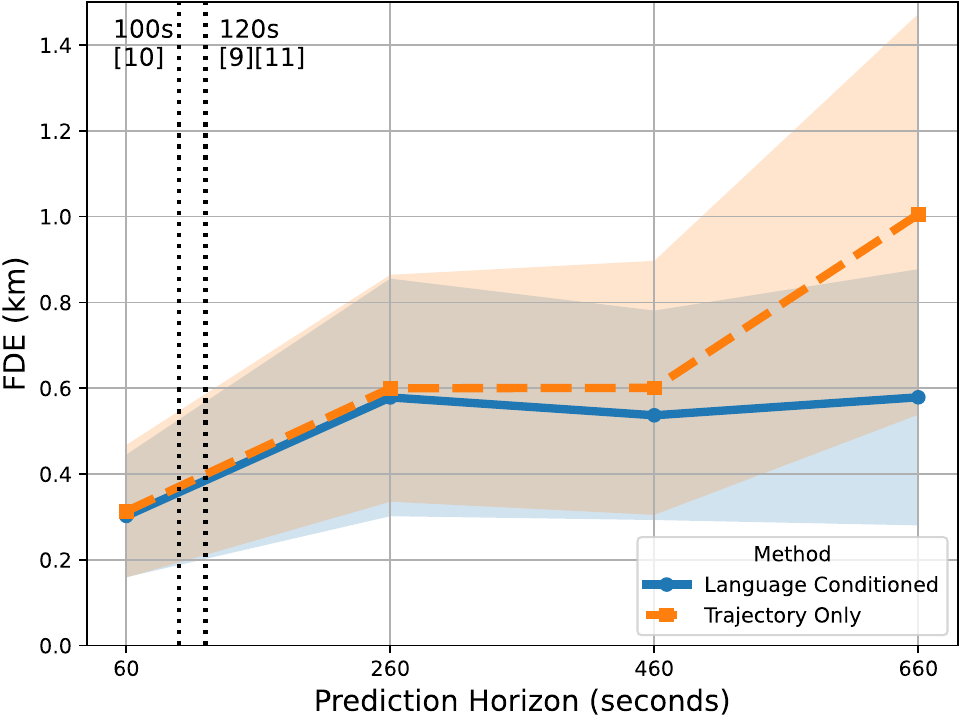}
    \caption{Final displacement error (FDE) of predicted aircraft goals vs. prediction time horizon length.
    Error increases with longer horizons, highlighting the challenge of long-term forecasting. However, language-conditioned goal prediction is better suited to address this challenge, showing a smaller increase in error. The prediction horizons used in TrajAirNet \cite{patrikar2022predicting}, GooDFlight \cite{yang2025goodflight} and Social-PatteRNN-ATT \cite{navarro2022social} are shown as vertical lines. IQR is plotted as a measure of dispersion.}
    \label{fig:pred_horizon}
\end{figure}

As shown in \cref{fig:obs_horizon}, the goal prediction error (FDE) of our model does not change significantly with an increased observation time horizon.
This result suggests that future goals are not correlated with the past position history, but rather with the current position and future intent.
In \cref{fig:pred_horizon}, however, goal prediction performance does change with the length of the prediction horizon.
While both the language conditioned model and trajectory-only model experience loss in accuracy when trying to predict longer horizon goals, this loss is greater for the trajectory-only model.

\begin{figure}[htbp]
    \centering
    \includegraphics[width=0.95\linewidth]{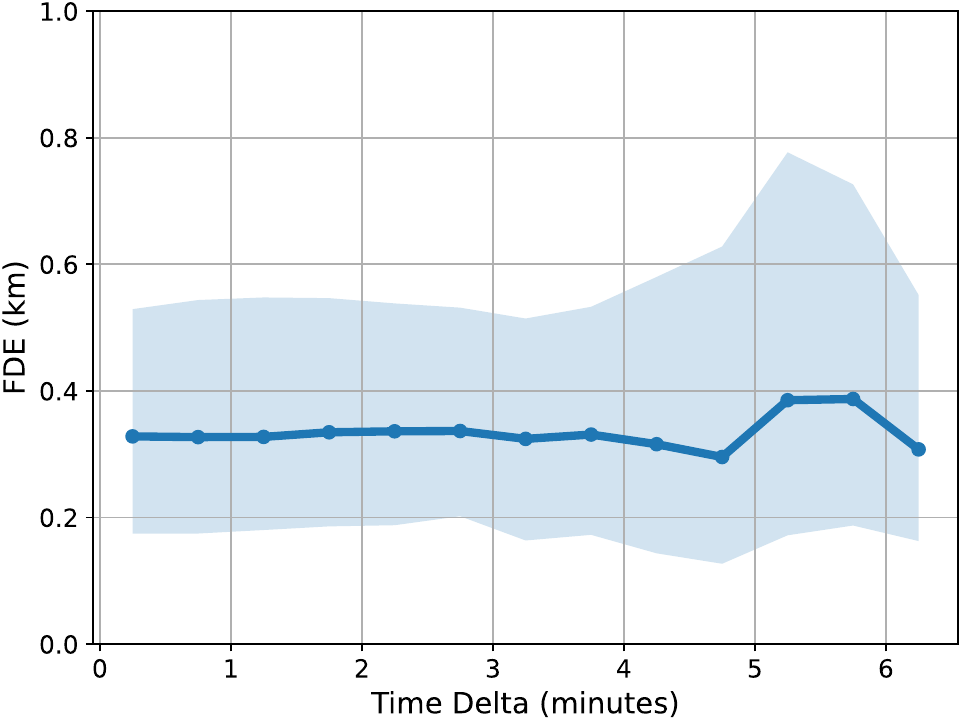}
    \caption{Final displacement error (FDE) of predicted aircraft goals vs. the time difference between the most recent radio call used for prediction and the time at which prediction is made.
    Our model does not exhibit significantly increased error with longer delay. IQR is plotted as a measure of dispersion.}
    \label{fig:time_delta}
    \vspace*{-12pt}
\end{figure}

Finally, the impact of the recency of radio calls is shown in \cref{fig:time_delta}.
Since radio calls are less frequently received than ADS-B position data, we condition predictions on radio calls that are up to 10 minutes old.
However, we see that using older radio calls does not result in significantly worse prediction performance, supporting the idea that radio calls contain long horizon intentions which persist over time.

However, for radio calls with ``Unknown'' intent labels, our model predicts goal positions with an FDE of  $0.5832\pm0.6544$ km whereas for all other intent labels the FDE is $0.4836\pm0.5131$ km.
This result suggests that the prediction performance is worse without intent information.

\section{Conclusion}\label{sec:conclusion}

The key finding of this paper is that conditioning on natural language radio calls can improve predictions of aircraft desired goal locations.
To achieve this, we proposed a new deep neural network model to approximate a distribution over possible goal positions based on an aircraft's recent trajectory history, its most recent radio call, and domain context.
Experiments showed that such conditioning on language improves goal prediction accuracy over baselines; we further confirm that ablating away language significantly decreases accuracy, and language conditioning improves prediction accuracy further into the future.
This improved prediction accuracy is a critical step towards satisfying proposed safety standards for autonomous aircraft, which stipulate a minimum separation distance of 1,500 feet (0.457 km) in terminal airspace \cite{rtca_do365_2017}.

Future work will address several key limitations.
First, we have not integrated goal prediction into a closed-loop autonomous pipeline.
Furthermore, our approach is specific to a single airport (which is a challenge shared by all of the baselines).
All baselines also currently share large variance in their predictions.
Finally, it remains open how best to autonomously \textit{generate} correct, informative radio calls.

\renewcommand{\bibfont}{\normalfont\footnotesize}
{\renewcommand{\markboth}[2]{}
\printbibliography}
\begin{appendices}
\crefalias{section}{appendix}
\crefalias{subsection}{appendix}
\section{}\label{appx:domain_context}
Example of dynamic context, the time-varying part of the domain context $d$, used to identify the transmitting aircraft from a radio call:

\textsf{\small Your task is to identify which aircraft made the call from a set of options. Only reply with the aircraft tail number with no additional text, if you cannot make a determination return Unknown.
Options:
N17NA - P28A, Piper, Carioquinha, Cherokee, Liner, 140-4, Chief, Archer, Cruiser, Challenger, Cadet, Warrior, Flite, Tupi
Location - 5 miles, North; 
N2423U - C172, Cutlass, Hawk, Rocket, Skyhawk, Mescalero, Reims, Powermatic
Location - 3 miles, East
}
\end{appendices}

\end{document}